%% file: sigasia_main_arxiv.tex
\PassOptionsToPackage{table,dvipsnames,x11names}{xcolor}
\documentclass[acmtog]{acmart}
\acmSubmissionID{2312}

\input{preamble} 
\acmJournal{TOG}

\definecolor{NavyBlue}{RGB}{0,0,128} 

\title{\textcolor{NavyBlue}{\textbf{ABACUS}}: \textcolor{NavyBlue}{\textbf{A}}dapting Unified Foundation Model for \textcolor{NavyBlue}{\textbf{B}}ridging Im\textcolor{NavyBlue}{\textbf{a}}ge \\ \textcolor{NavyBlue}{\textbf{C}}ount \textcolor{NavyBlue}{\textbf{U}}nder\textcolor{NavyBlue}{\textbf{s}}tanding and Generation}

\author{Anindya Mondal*}\thanks{*Authors have equal contributions.}
\affiliation{
  \institution{University of Surrey}
  \country{United Kingdom}}
\email{a.mondal@surrey.ac.uk}

\author{Sauradip Nag*}
\affiliation{
  \institution{University of Surrey, United Kingdom \& Simon Fraser University}
  \country{Canada}
}
\email{sauradipnag95@gmail.com}

\author{Anjan Dutta}
\affiliation{
 \institution{University of Surrey}
 \country{United Kingdom}}
 \email{anjan.dutta@surrey.ac.uk}

\renewcommand{\shortauthors}{Mondal et al.}

\begin{document}

\begin{abstract}
\input{final_arxiv/00_abs}
\end{abstract}

\begin{teaserfigure}
    \centering
    \includegraphics[width=\linewidth]{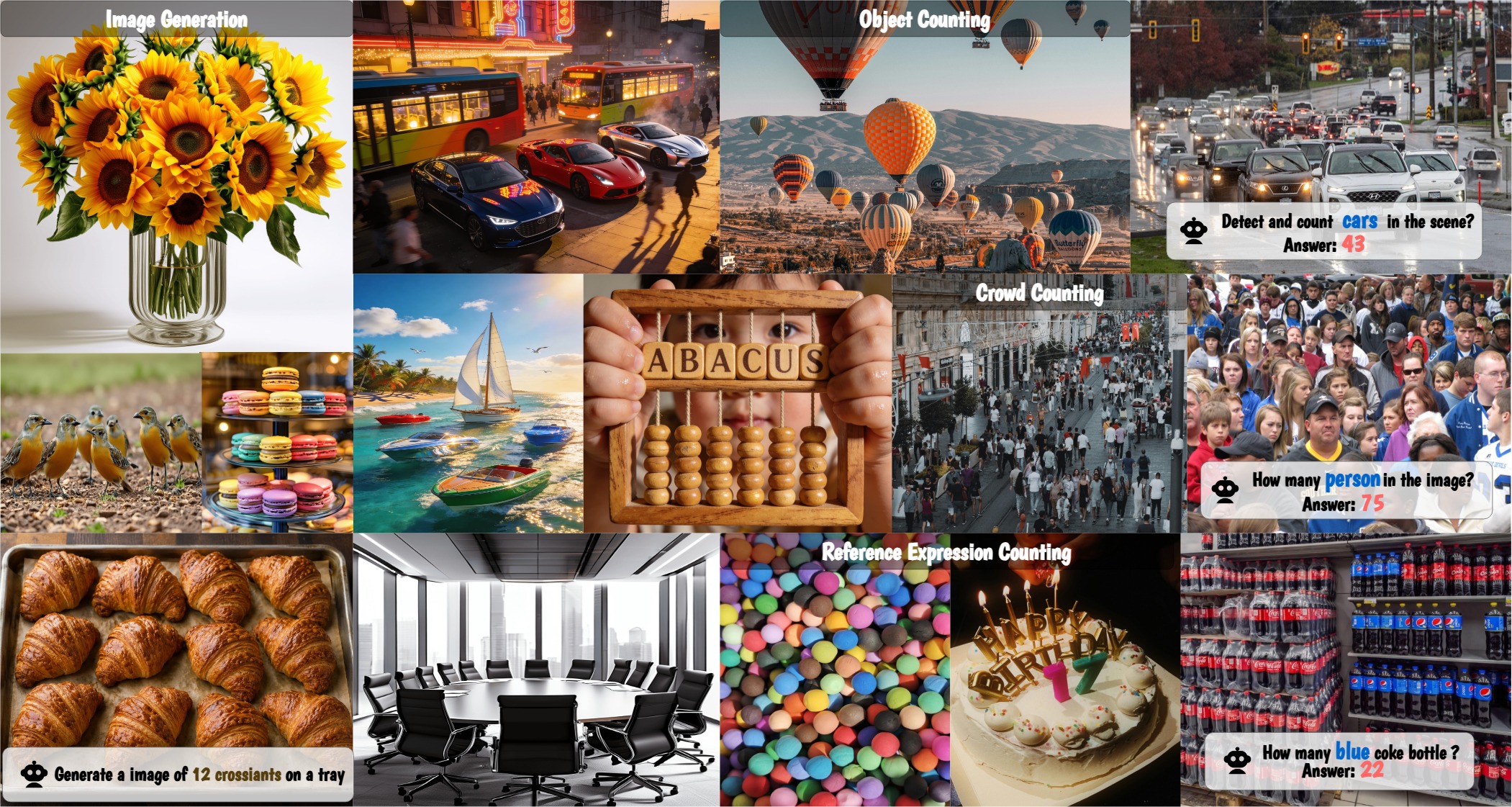}
    \caption{\change{\textbf{\model{} overview.} A single unified model performs count-aware image generation (left), object counting, crowd counting, and referring-expression counting (right) using text-only prompts with no benchmark-specific training. }}
    \label{fig:teaser}
\end{teaserfigure}

\maketitle

\input{final_arxiv/01_intro}
\input{final_arxiv/02_lit}
\input{final_arxiv/03_model}
\input{final_arxiv/04_exp}
\input{final_arxiv/05_conc}

\newpage
\bibliographystyle{ACM-Reference-Format}
\bibliography{references}

\newpage

\end{document}

%% file: preamble.tex
\usepackage{graphicx,amsmath,amsthm,booktabs,xcolor,siunitx,microtype,nicefrac,enumitem,amsfonts,tikz}
\newcolumntype{d}[1]{S[table-format=#1]}
\definecolor{band}{RGB}{245,248,252} 

\usepackage{listings}
\usepackage[ruled,vlined,linesnumbered]{algorithm2e}
\usepackage{amsmath}
\usepackage{multirow}
\usepackage{dsfont}
\usepackage{cutwin}
\usepackage{mdframed}
\usepackage{pifont}
\usepackage[utf8]{inputenc} 
\usepackage[T1]{fontenc}
\usepackage{gensymb}
\usepackage{textcomp}
\usepackage{float} %
\usepackage{cleveref}
\usepackage[many]{tcolorbox} 
\tcbuselibrary{skins}     
\tcbuselibrary{breakable}  
\newmdenv[linewidth=0.8pt,leftmargin=0pt,topline=false,rightline=false,bottomline=false]{assbox}

\SetAlFnt{\small}
\SetAlCapFnt{\small}
\SetAlCapNameFnt{\small}
\SetAlCapHSkip{0pt}

\newcommand{\change}[1]{\textcolor{black}{#1}}

\usepackage[table,x11names]{xcolor} \definecolor{T2IBlue}{HTML}{E8F0FE}      
\definecolor{L2IGreen}{HTML}{E6F4EA}     
\definecolor{AgenticPurple}{HTML}{EFE8F4}  
\definecolor{OursGray}{HTML}{F5F5F5}     
\definecolor{ablationBG}{HTML}{F3F7FF} 
\definecolor{criticBG}{HTML}{FFF7F0}   
\definecolor{userBG}{HTML}{EEF8EF}     

\usepackage{colortbl}
\definecolor{secbg}{RGB}{220,230,242}      
\definecolor{ourbg}{RGB}{255,247,215}      
\definecolor{ruleblue}{RGB}{90,120,160}    
\usepackage{color,soul}
\usepackage[font=small,labelfont=bf]{caption} 
\usepackage{subcaption}                        
\usepackage{amsmath}

\usepackage{amssymb}
\usepackage{cleveref}
\usepackage{newfloat}
\DeclareCaptionStyle{ruled}{labelfont=normalfont,labelsep=colon,strut=off} 
\newcommand{\eg}{\textit{e.g.}}

\def\model{\textsc{ABACUS}}
\newcommand{\myparagraph}[1]{\vspace{0pt}\noindent{\bf #1:}}
\definecolor{DarkGreen}{RGB}{47,198,0}
\newcommand{\cmark}{\ding{51}}%

%% file: final_arxiv/00_abs.tex
We present ABACUS, a unified vision-language model that jointly addresses object counting, crowd counting, referring-expression counting, and count-faithful image generation within a single 3B-parameter {model}. ABACUS introduces three contributions: density-aware adaptive zooming paired with an objectness map from multi-head self-attention decomposition to spatially ground count predictions; a boundary-aware count policy trained via GRPO with nested local, boundary, and global rewards to eliminate over- and undercounting at crop boundaries; and a cycle-consistent GRPO strategy in which the frozen understanding branch scores generated candidates on count-deviation and aesthetic quality, closing the understanding–generation synergy gap without any external critic or annotation. ABACUS achieves state-of-the-art results across seven benchmarks spanning object counting (FSC-147, CARPK), crowd counting (ShanghaiTech A/B), referring-expression counting (REC-8K), count-faithful generation (CoCoCount, T2I-CompBench, GenEval), and count reasoning (CountQA), surpassing both task-specific specialists and larger generalist models. Project page is at \textcolor{red}{\href{https://mondalanindya.github.io/pages/ABACUS/}{https://mondalanindya.github.io/pages/ABACUS/}}.


%% file: final_arxiv/01_intro.tex
\section{Introduction}

\begin{figure}[t]
    \centering
    \includegraphics[width=\linewidth]{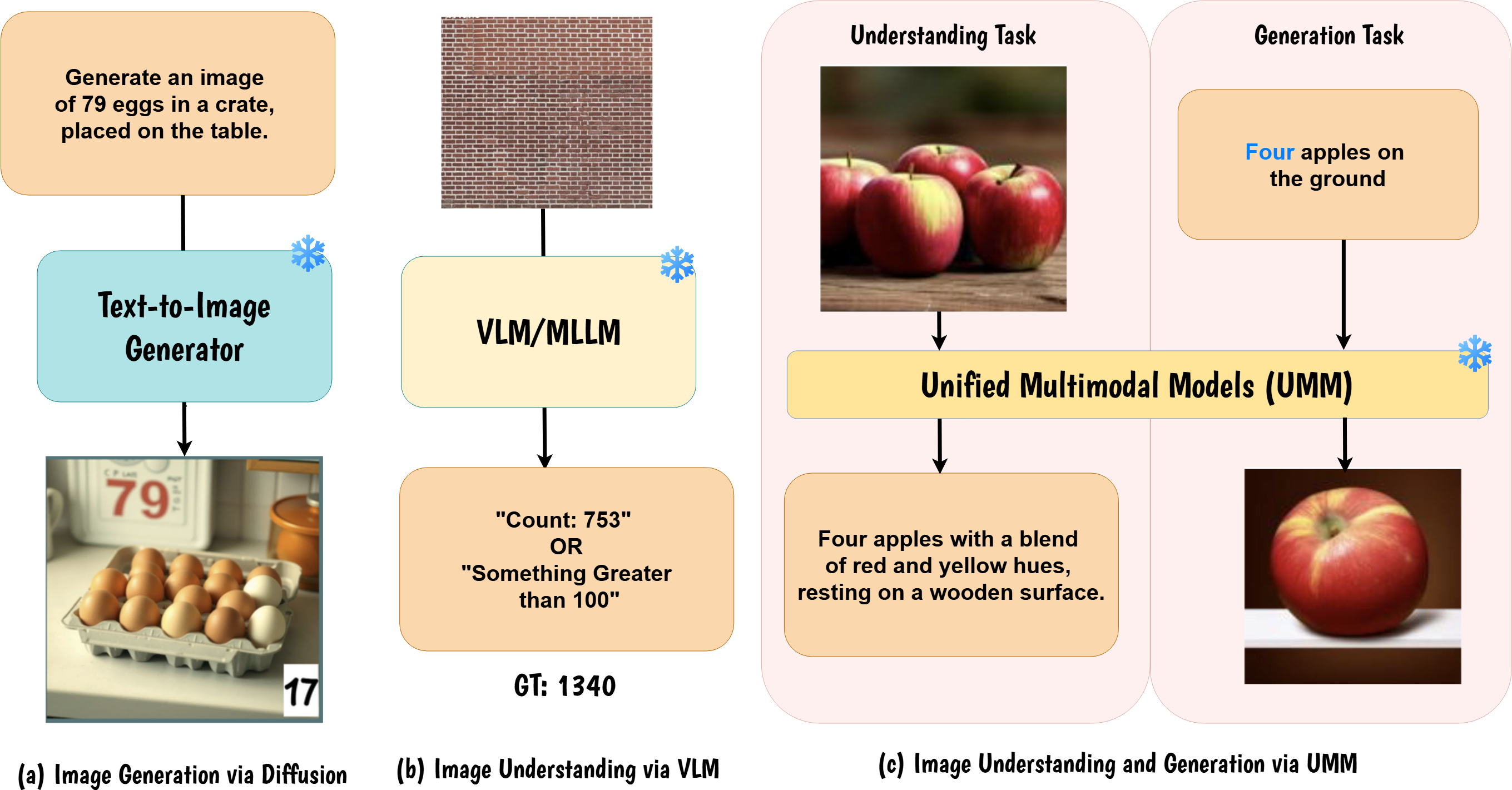}
    \caption{\change{
    \textbf{Issues in Count Generation and Understanding.} 
(a)~Text-to-image diffusion models lack any mechanism to verify output 
cardinality. (b)~VLMs and MLLMs default to coarse magnitude estimates 
(\eg, ``Greater than~100'') on dense scenes. (c)~Existing UMMs support 
both tasks from a single models yet exhibit a \emph{synergy gap}: 
the same model that correctly counts 4~apples cannot generate exactly 4. 
}}
    \label{fig:issues}
\end{figure}

Object counting~\cite{ranjan2021learning,liu2022countr,AminiNaieni23} is the 
task of estimating the number of target instances in an image, conventionally 
addressed by regressing a density map whose spatial integral gives the total 
count. Count-conditioned image generation~\cite{binyamin2025make,
kang2023countingguidance} is the complementary task of synthesising a scene 
containing precisely the number of instances specified in a text prompt. Despite 
remarkable progress in both directions, these tasks have been pursued in 
isolation — object counting~\cite{ranjan2021learning,amini2023countx,
amini2025countgd++}, crowd counting~\cite{shu2022crowd,liang2023crowdclip}, 
referring-expression counting~\cite{dai2024referring}, and count-conditioned 
generation~\cite{binyamin2025make,kang2023countingguidance,mondal2025countloop} 
each rely on task-specific architectures, loss formulations, and training 
pipelines that do not generalise across regimes. Attempts to unify these under 
broader vision-language models~\cite{amini2025countgd++,zhang2026wscoc} have 
shown limited success: such models generalise across object categories but 
collapse on fine-grained, instance-level instructions due to a lack of spatial 
grounding during training. At the heart of this difficulty lies \emph{objectness} 
 the coherent representation of a distinct entity even amid visually similar 
neighbours~\cite{alexe2012measuring,kuo2015deepbox} a capacity long studied 
in cognitive psychology~\cite{spelke1990principles} and still poorly captured 
by current models. On the generation side, state-of-the-art diffusion 
models~\cite{binyamin2025make,dahary2024yourself} routinely produce incorrect 
cardinalities (see Fig~\ref{fig:issues}(a)) or spatially incoherent arrangements, underscoring that generating 
a precise number of objects requires reasoning about global spatial relations 
that purely generative objectives do not enforce. These limitations motivate a 
unified model capable of jointly performing count understanding and count-faithful 
generation from a shared representation, without task-specific supervision.

Recent unified multimodal models~\cite{xie2024show,chen2025janus,tang2025unilip,
deng2025bagel} have demonstrated that visual understanding and generation can 
be handled within a shared parameter space, yet closing the gap between the two 
capabilities remains non-trivial. Existing approaches rely on carefully engineered 
training recipes — including data and loss balancing, multi-stage optimisation, 
and hybrid MLLM-diffusion pipelines — to prevent gains in one task from 
degrading the other. Even under such regimes, a persistent \emph{synergy gap} 
remains: as illustrated in \cref{fig:issues}(c), the same unified checkpoint that 
correctly counts four apples in an image fails to generate exactly four apples 
from a text prompt. This asymmetry is compounded by a more fundamental 
deficiency, off-the-shelf MLLMs struggle to parse dense visual 
scenes~\cite{qharabagh2024lvlm} (Fig~\ref{fig:issues}(a)), so the understanding branch of native 
UMMs cannot provide reliable count supervision for the generation branch. 
Consequently, neither task benefits from the complementary signal the other 
could in principle offer.


To bridge this gap, we introduce \textbf{ABACUS}, a unified vision-language 
model built upon~\cite{tang2025unilip} that jointly addresses object counting, 
crowd counting, referring-expression counting, and count-faithful image 
generation within a single unified model, generalising to real-world scenarios 
in a zero-shot manner. To overcome the well-documented failure of MLLMs on dense scenes~\citep{qharabagh2024lvlm,zhang2026wscoc}, we propose 
density-aware adaptive zooming, which recursively partitions dense images into 
manageable sub-regions so that the MLLM operates over locally sparse fields 
where per-instance discrimination is more reliable for counting. Complementing this, we 
derive an objectness map from multi-head self-attention 
decomposition~\cite{vaswani2017attention} of the MLLM attention layers that spatially grounds count 
predictions in genuine per-instance evidence, steering the model away from  spurious token memorisation~\cite{chu2025sft} towards principled spatial  reasoning. Since recursive partitioning inevitably places objects at crop  boundaries, we further introduce a boundary-aware count policy trained via  GRPO~\cite{shao2024deepseekmath} as a post-training objective, which  explicitly resolves ownership of straddling instances through nested local,  boundary, and global rewards, yielding a substantially more accurate and consistent counter. With the understanding branch thus stabilised, we use it as a frozen 
counting model to improve the generation branch. Since the two tasks are  naturally complementary:  understanding maps images to text while generation maps text to images, the understanding branch can directly  evaluate how well a generated image matches its text prompt, providing 
supervision without any external critic. Concretely, we adopt a  cycle-consistent GRPO strategy~\cite{shao2024deepseekmath} where the 
generation branch produces a group of candidate images for each  count-conditioned prompt; the frozen understanding branch scores each candidate based on how well it matches the requested count; an external aesthetic scorer provides a complementary image quality  reward; and the combined rewards update only the generation branch.  This closed loop progressively improves count-faithful generation  without any external model, or human 
annotation.

In summary, our contributions are as follows:

\begin{itemize}[leftmargin=*,itemsep=2pt,topsep=2pt]

\item We present \model{}, the first unified VLM that jointly addresses all count understanding tasks like Object counting, Crowd counting, Reference Expression counting and count-accurate image generation in a zero-shot manner.

\item We introduce density-aware adaptive zooming paired with an objectness map obtained from MLLM attention layers to  spatially ground count predictions, and a novel boundary-aware count policy via GRPO to eliminate the over/undercounting artifact introduced at crop boundaries.

\item We propose a cycle-consistent GRPO strategy that uses the frozen understanding branch as an ideal counting model to score generated images via  count-deviation and aesthetic rewards, updating only the generation 
branch to close the understanding--generation synergy gap.

\item \model{} sets a new state of the art across seven benchmarks spanning object counting, crowd counting, referring-expression counting, count-faithful generation, and count reasoning, surpassing both task-specific specialists and larger generalist models with a single 3B-parameter checkpoint.

\end{itemize}

%% file: final_arxiv/02_lit.tex
\section{Related work}

\subsection{Count Understanding}

Existing counting methods are broadly divided into class-specific 
and class-agnostic approaches. Class-specific methods predict counts 
for fixed categories such as persons~\cite{ranasinghe2024crowddiff,
guo2024regressor,shi2019revisiting}, vehicles~\cite{hsieh2017drone,
mundhenk2016large}, and cells~\cite{xie2018microscopy} via 
detection~\cite{liu2023point,liang2022end} or density-map 
regression~\cite{wang2020distribution,li2023calibrating}. 
Class-agnostic methods~\cite{chattopadhyay2017counting,lu2019class,
Xu_2023_CVPR} accept visual exemplars~\cite{liu2022countr,
pelhan2024dave} or text prompts~\cite{AminiNaieni24,kang2024vlcounter,
qian2025t2icount} as targets, though most require dense point-level 
supervision. Crowd counting~\cite{zhang2016sHAB,li2018csrnet,
wang2020distribution} focuses on dense person scenes via density-map 
regression or point-level detection~\cite{yin2018p2p,liu2023point}, 
with recent work addressing domain shift~\cite{du2023domain,
peng2024single}, uncertainty~\cite{li2023calibrating}, and weakly 
supervised generalisation~\cite{zhang2026wscoc}. Referring Expression 
Counting (REC)~\cite{Dai_2024_CVPR} further generalises 
class-agnostic counting to free-form expressions requiring joint 
reasoning over attributes, spatial relations, and category; to date, 
REC has been addressed exclusively by specialist detection-based 
models with box-level supervision~\cite{wang2025exploring}. 
\model{} unifies all three regimes — object counting, crowd 
counting, and referring-expression counting — within a single 
zero-shot checkpoint, estimating counts autoregressively via 
next-token prediction without density maps, counting heads, or 
point-level annotations, and reports the first unified-VLM result 
on REC-8K.

\subsection{Count Generation}
Text-to-image diffusion models~\cite{podell2023sdxl,betker2023improving,flux2024,alimohammadi2025cora,alimohammadi2025smite,vadakkeeveetil2026respodiff} persistently struggle with accurate object counting~\cite{paiss2023teaching}. Previous solutions typically rely on architecturally disjoint counting signals, such as contrastive losses~\cite{paiss2023teaching}, auxiliary heads~\cite{kang2023countingguidance}, or attention manipulation~\cite{chefer2023attendexcite}. While spatial planners~\cite{feng2023layoutgpt,li2023gligen,binyamin2025make} offer tighter integration, they often fail to generate all specified objects or produce rigid layouts at higher counts~\cite{dahary2024yourself,tewel2024training}. Furthermore, recent iterative approaches employing external Vision-Language Model (VLM) critics are inherently bottlenecked by critic reliability~\cite{mondal2025countloop,wallace2024diffusion}. To address these limitations, \model{} unifies the generator, counter, and verifier into a single architecture. By enabling the understanding branch to directly reward the generation branch during training, \model{} entirely eliminates the need for external critics, planners, or annotations.

\subsection{Multimodal Large Models}
Early vision-language models (VLMs) such as CLIP~\cite{radford2021learning} 
and BLIP~\cite{li2022blip} align vision and text through contrastive 
pretraining and have been widely adopted for downstream 
tasks~\cite{jiang2023clip,kang2024vlcounter,nag2018offline,nag2022zero,nag20252,ramakrishnan2025dreampet,nag2022multi,nag2023difftad,nag2022proposal}. More recently, 
multimodal large language models 
(MLLMs)~\cite{liu2024llavanext,li2024llava,Qwen2.5-VL} extend these 
capabilities with stronger reasoning and generation, achieving 
notable results on visual question answering~\cite{Goyal2017vqav2} 
and image captioning~\cite{agrawal2019nocaps}.
{Multimodal Large Language Models (MLLMs) often generate confident yet misaligned outputs \cite{huang2024opera,sun2024aligning}. While Reinforcement Learning from Human Feedback (RLHF) effectively mitigates this \cite{sun2024aligning, yu2024rlhf}, its reliance on costly manual annotation limits scalability. Consequently, Reinforcement Learning from AI Feedback (RLAIF) has emerged as a practical alternative \cite{lee2023rlaif,li2023silkie}. Within the counting domain, \cite{mondal2025countloop} leveraged AI feedback for count-conditional image generation, employing a critic VLM to iteratively refine outputs without computationally heavy RL training. Conversely, for count understanding, prior works have adapted VLMs for text-driven object counting~\cite{jiang2023clip,qharabagh2024lvlm}. Methods like CLIP-Count~\cite{jiang2023clip} and VLCounter~\cite{kang2024vlcounter} append a specialized counting head to a fine-tuned CLIP backbone. Although WS-COC~\cite{zhang2026wscoc} bypasses explicit heads by autoregressively predicting counts via an MLLM, it remains restricted to weakly-supervised, class-agnostic tasks. In contrast, \model{} unifies count understanding encompassing object, crowd, and referring-expression counting within a single zero-shot framework. Simultaneously, it enables count-accurate image generation, an unprecedented capability among VLM-based counting methods. We achieve this by fine-tuning our architecture via a novel RLAIF-driven policy optimization.}

%% file: final_arxiv/03_model.tex
\section{Method}
\label{sec:method}

We present \model, a framework that endows a single unified vision-language model (VLM) with both count-accurate image \emph{generation} and precise object \emph{understanding} (counting). {We first discuss the preliminaries for RL based model training (Sec~\ref{sec:grpo_prelim}), then we talk about how ABACUS counts object in an image (Sec~\ref{sec:under}), then we discuss how we can generate high cardinality images using the same model (Sec~\ref{sec:count_gen}) and finally, we discuss about the training strategies (Sec~\ref{sec:train}) respectively.}

{\subsection{Preliminaries: Group Relative Policy Optimisation}
\label{sec:grpo_prelim}
{
\model{} uses Group Relative Policy Optimisation (GRPO)~\cite{shao2024deepseekmath} as a unified post-training mechanism for both the understanding and generation branches.
Unlike PPO, GRPO eliminates the value network by computing advantages relative to a group of rollouts from the policy itself, reducing memory and compute overhead.}
{Given a policy $\pi_\theta$, an input context $\mathbf{x}$, and a task-specific scalar reward function $R(\cdot)$, GRPO samples $K$ rollouts $\{\mathbf{y}_i\}_{i=1}^{K} \sim \pi_\theta(\cdot \mid \mathbf{x})$ and computes group-relative advantages:}
\begin{equation}
    A_i \;=\;
    \frac{R(\mathbf{y}_i) - \mu_R}{\sigma_R + \epsilon},
    \qquad
    \mu_R = \frac{1}{K}\sum_{j=1}^{K} R(\mathbf{y}_j),
    \quad
    \sigma_R = \mathrm{std}\bigl(\{R(\mathbf{y}_j)\}_{j=1}^{K}\bigr).
    \label{eq:advantage}
\end{equation}
The policy is updated by maximising the surrogate objective with a KL penalty against a frozen reference policy $\pi_{\mathrm{ref}}$:
\begin{equation}
    \mathcal{L}_{\mathrm{GRPO}}(\theta) \;=\;
    \mathbb{E}\!\left[
        \sum_{i=1}^{K} A_i \log \pi_\theta\!\left(
            \mathbf{y}_i \mid \mathbf{x}
        \right)
    \right]
    \;-\; \beta\,\mathrm{D}_{\mathrm{KL}}\!\bigl[
        \pi_\theta \;\|\; \pi_{\mathrm{ref}}
    \bigr],
    \label{eq:grpo}
\end{equation}
{where $\beta > 0$ controls regularisation strength.
In \cref{sec:method}, we instantiate this framework twice with different reward functions: a boundary-aware counting reward for the understanding branch (\cref{sec:under}) and a count-deviation self-reward for the generation branch (\cref{sec:count_gen}).}}

\subsection{Count Understanding}
\label{sec:under}
Predicting absolute object counts is challenging for MLLMs, 
particularly in dense scenes without per-instance supervision. 
We address this by adaptively partitioning input images into 
sparser sub-regions, allowing the MLLM to produce more reliable 
local count estimates that are then aggregated into a global count.

\begin{figure}[t]
    \centering
    \includegraphics[width=\linewidth]{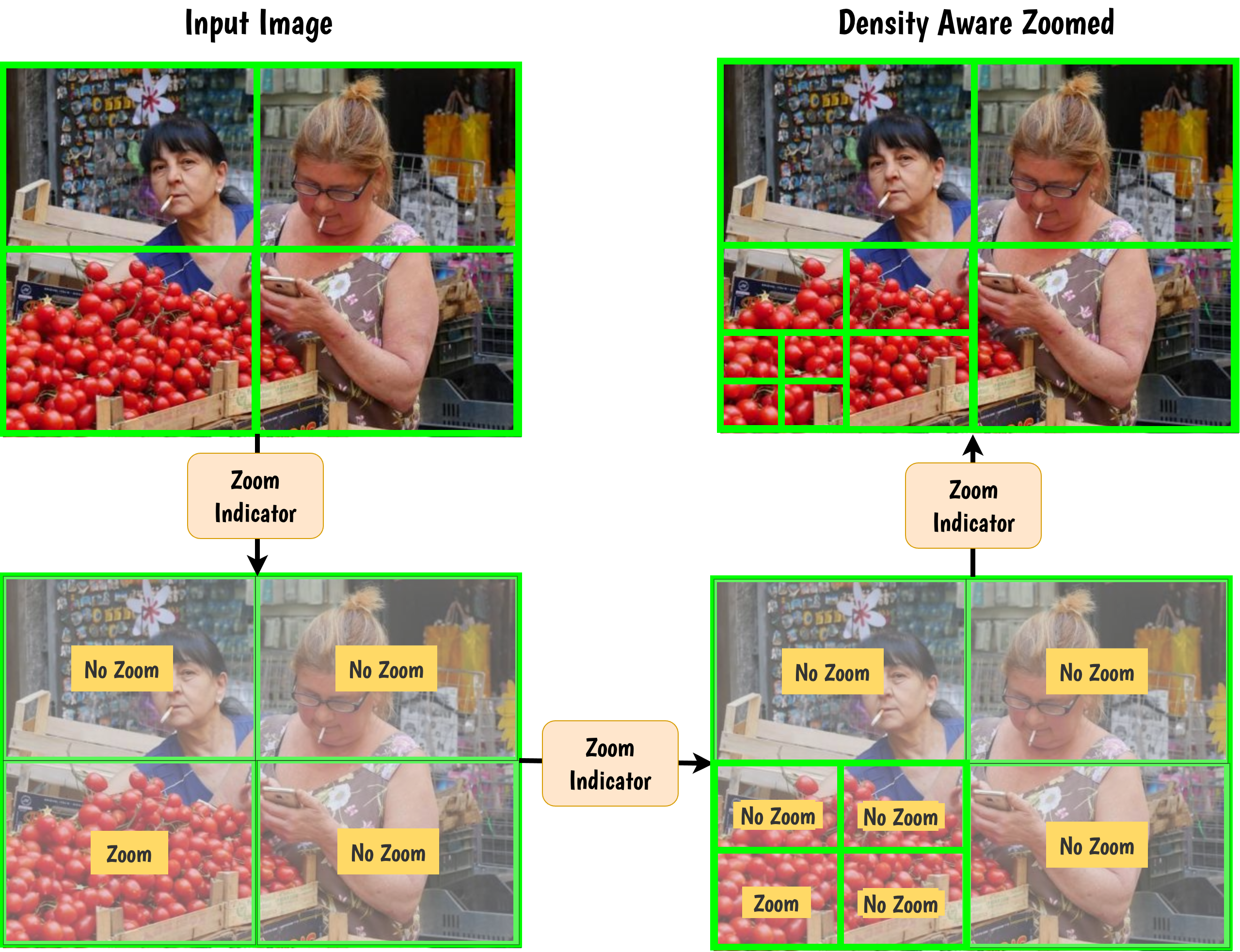}
    \caption{\change{\textbf{Density-aware adaptive zooming.} The zoom indicator $\phi(\cdot)$ classifies each image region as sparse or dense. Dense regions are recursively partitioned into $2{\times}2$ sub-regions until resolution $\gamma$ is reached; sparse regions are processed in a single pass. Local predictions are aggregated to produce the global count.}}
    \label{fig:placeholder}
\end{figure}

\noindent \textbf{Density-aware Adaptive Zooming}
Since MLLMs are more reliable on sparse regions, we recursively 
partition dense images into sub-regions with lower local counts. 
Partitioning is governed by a zoom indicator $\phi(\cdot)$, 
implemented as a frozen GroundingDINO~\cite{liu2024gdino} backbone 
with a lightweight learnable MLP head trained on a curated 
sparse/dense binary dataset. Given an input image $I$, the indicator 
computes a density score $s_d$; if the image is classified as dense, 
it is recursively split into $2\times2$ sub-regions until a minimum 
resolution $\gamma$ is reached:
\begin{equation}
    \{I_{i},\ldots,I_{n}\} = \phi(I;\gamma;s_d), \quad 
    C_{i} = \mathbb{M}_{\theta}(I_{i};\mathcal{T}),
    \label{eq:zoom}
\end{equation}
where $\mathbb{M}_{\theta}(\cdot)$ denotes the MLLM head with 
learnable parameters $\theta$. Following~\cite{tang2025unilip}, we 
append learnable meta-tokens to the MLLM input. Each sub-image 
$I_{i}$ is queried independently to estimate the count of the 
target specified in text prompt $\mathcal{T}$, and the local 
predictions $C_{i}$ are aggregated to produce the final global count.

\noindent \textbf{Infusing Objectness in MLLM} 
Although MLLMs exhibit strong reasoning, they struggle with precise 
object localisation~\cite{chen2025revisiting}. Rather than relying 
on explicit bounding-box prompting, which frequently produces 
hallucinated coordinates for small objects~\cite{kazemzadeh2014referitgame}, 
we extract spatial object evidence directly from the MLLM's internal 
representations. Recent work shows that visual tokens largely 
preserve spatial correspondence to their originating image regions 
across transformer layers~\cite{neo2025towards,vaswani2017attention}; 
we exploit this by decomposing multi-head self-attention (MHSA) to 
derive a per-patch objectness signal.
Concretely, for each layer $l$ and head $i$, we isolate the head's 
contribution via structured masking over the pre-projection 
concatenated outputs, pass it through the frozen output projection 
$W_O$, and apply a shared learned affine alignment 
$\mathcal{A}(\cdot)$~\cite{belrose2023tuned} to obtain a 
geometrically consistent residual $\mathbf{r}^{l,i}$. The objectness 
score at each visual token position $v$ is its $\ell_2$ magnitude:
\begin{equation}
    o^{l,i}(v) = \left\|\mathbf{r}^{l,i}_v\right\|_2, \quad v \in \mathcal{V}.
    \label{eq:objectness}
\end{equation}
A spatial attention distribution $q^l(v)$ is further derived by 
averaging cross-attention weights from the final generative token to 
all visual positions across heads. The objectness map is supervised 
against Gaussian-smoothed ground-truth point annotations via a 
objectness regularisation loss:
\begin{equation}
    \mathcal{L}_{\text{obj}} = 
    \frac{1}{|\mathcal{L}|} \sum_{l \in \mathcal{L}} 
    \left(-\sum_{v \in \mathcal{V}} g(v)\log\!\left(\tilde{q}^{l}(v) 
    + \epsilon\right)\right),
    \label{eq:focus_loss}
\end{equation}
where {$g(v)$ denotes the Gaussian-smoothed ground-truth
target, obtained by centring a 2D Gaussian kernel ($\sigma = 20$\,px) at
each annotated point location and rendering the resulting heatmap at the
MLLM attention-map resolution, and $\tilde{q}^{l}(v)$ is the binarised
predicted objectness map. {Smoothing converts sparse
point supervision into a spatially extended target that matches the
continuous space of the attention distribution; aligning the
objectness map with these targets therefore} encourages the MLLM to
internalise instance-aware spatial structure, steering count predictions
away from spurious token memorisation and towards principled
per-instance spatial reasoning.}


\begin{figure}[!t]
    \centering
    \includegraphics[width=\linewidth]{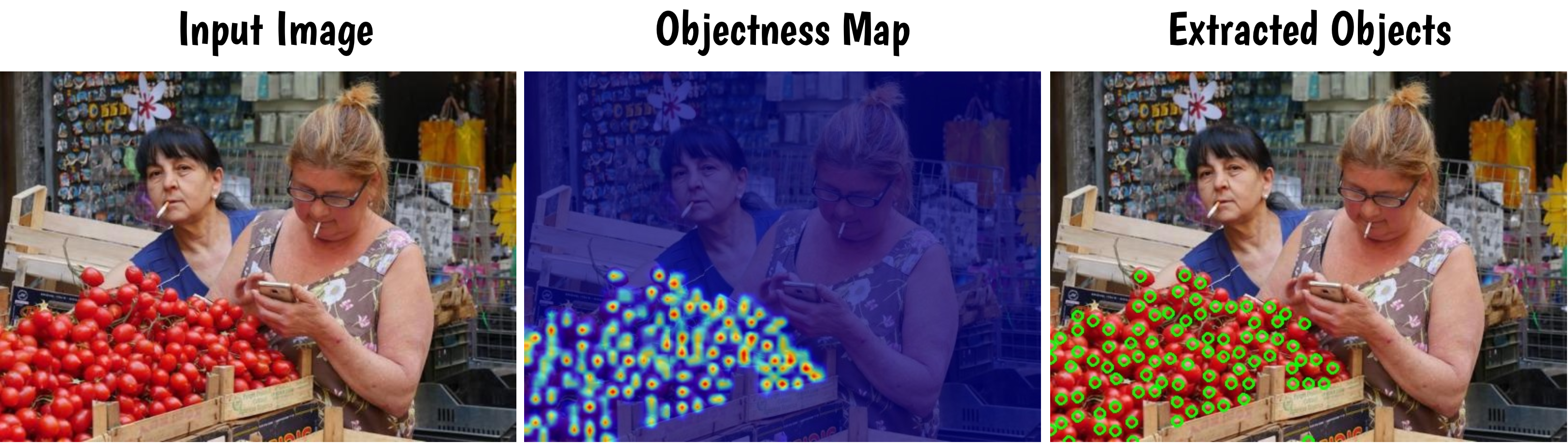}
    \caption{\change{\textbf{Infusing objectness in the MLLM.} The objectness map is extracted from the attention layers of the language model. Per-head isolation and learned affine alignment produce a spatial distribution over visual token positions, from which object peaks are identified. 
    }}
    \label{fig:placeholder}

\end{figure}

\noindent\textbf{Boundary-Aware Count Policy} 
Adaptive zooming inevitably places objects at crop boundaries, 
causing systematic over- or undercounting~\cite{qharabagh2024lvlm,
zhang2026wscoc}. To resolve this, we introduce a novel boundary-aware 
count policy that trains the MLLM to explicitly reason about 
boundary ownership. Given a dense image partitioned into a 
$2{\times}2$ non-overlapping grid of quadrant crops 
$\{Q_q\}_{q\in\mathcal{Q}}$, where $\mathcal{Q} = \{\text{TL}, 
\text{TR}, \text{BL}, \text{BR}\}$, the policy $\pi_\theta$ 
produces a structured output $\mathbf{y} = \pi_\theta(\{Q_q\}, 
\mathcal{P})$ classifying each object as \emph{interior} (interior 
count $n_q^{\text{int}}$, centroid within $Q_q$), \emph{edge} 
(edge-count $c_q^e$, centroid on this side of cut edge 
$e \in \mathcal{E}_q$), or \emph{boundary} (boundary-count $d_q^e$, 
centroid in the adjacent quadrant). Objects falling exactly on a 
crop line are assigned to one quadrant randomly, making 
double-counting structurally impossible. The per-quadrant subtotal 
$s_q$ and predicted global count $\hat{T}$ are:
\begin{equation}
    s_q = n_q^{\text{int}} + \sum_{e\,\in\,\mathcal{E}_q} c_q^e,
    \qquad
    \hat{T} = \sum_{q\,\in\,\mathcal{Q}} s_q.
    \label{eq:subtotal}
\end{equation}
The policy is post-trained via GRPO~\cite{shao2024deepseekmath} 
using three nested reward components at different spatial 
granularities: per-quadrant local accuracy ($\Delta^{q}_{r}$), 
cross-quadrant boundary consistency ($\Delta^{b}_{r}$), and global 
count coherence ($\Delta^{g}_{r}$), each computed as:
\begin{equation}
    \Delta_{r} = \exp\!\left(-\frac{|\text{pred} - \text{GT}|}
    {\text{GT}+\epsilon}\right),
    \label{eq:reward}
\end{equation}
which softly penalises over- and undercounting while remaining differentiable, {$\epsilon = 1e^{-6}$} preventing division by zero. For each training image, $K$ rollouts 
$\{\mathbf{y}_i\}_{i=1}^{K}$ are sampled from $\pi_\theta$; 
group-relative advantages are computed as 
$A_i = (R(\mathbf{y}_i) - \mu_R) / (\sigma_R + \epsilon)$, 
where $\mu_R$ and $\sigma_R$ are the mean and standard deviation 
of rewards within the group. The policy is then updated by 
maximising $ \mathcal{L}_{\mathrm{GRPO}}(\theta)$ (using Eq~\ref{eq:grpo}). 
%
%
where $\pi_{\mathrm{ref}}$ is a frozen reference policy 
(the fine-tuned understanding branch checkpoint) and $\beta > 0$ 
controls the KL penalty strength against reward hacking. The nested 
rewards jointly drive the policy to resolve boundary ambiguity at 
every spatial granularity, eliminating the systematic counting 
errors introduced by adaptive zooming.

\begin{figure}[!t]
    \centering
    \includegraphics[width=\linewidth]{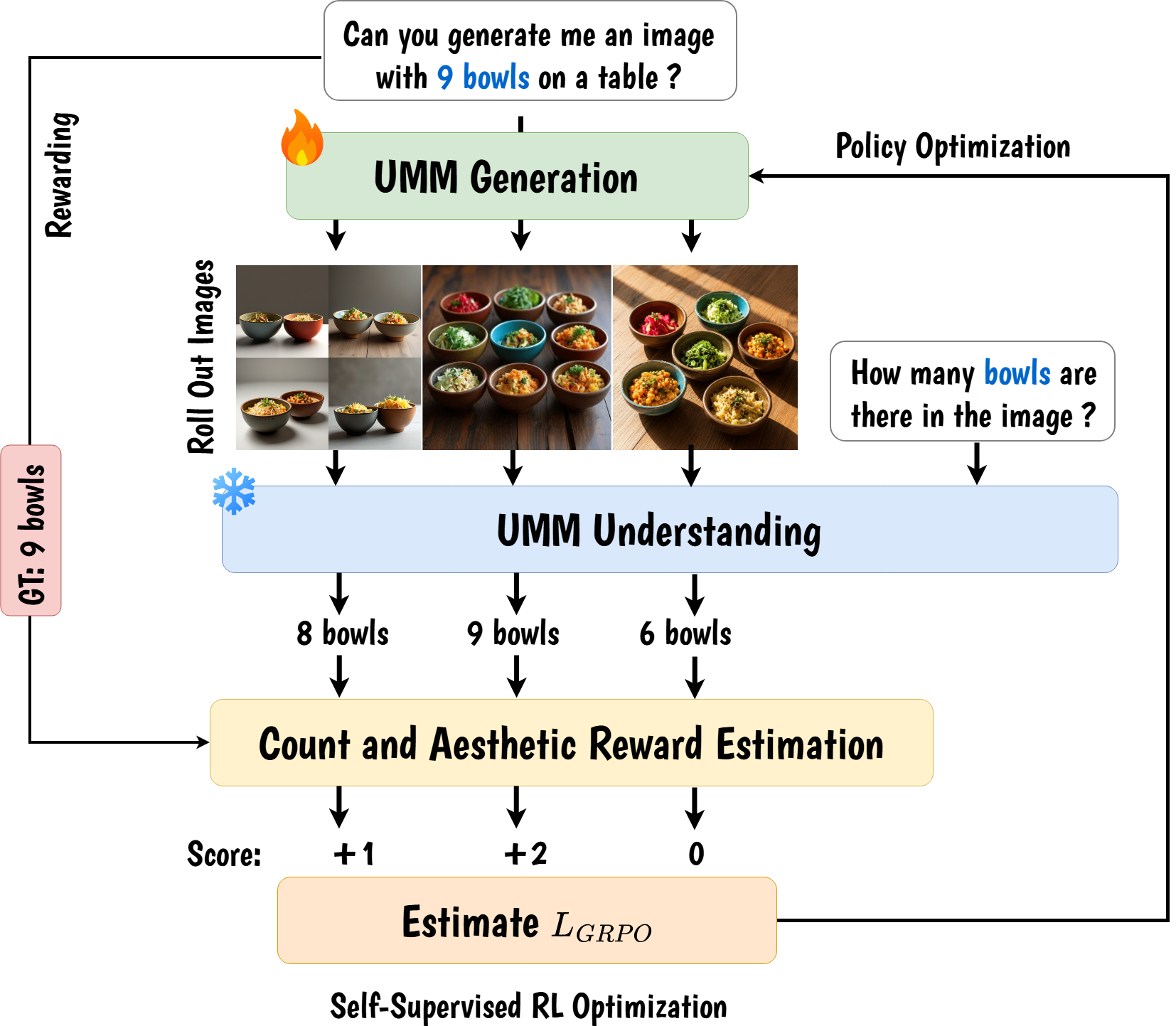}
    \caption{\change{\textbf{Count-aware image enhancement via cycle-consistent GRPO.} The generation branch samples $N$ candidate images from a count-conditioned prompt. The understanding branch counts objects in each candidate and scores aesthetic quality. The count deviation and aesthetic score form the GRPO reward, which updates the generation branch.}}
    \label{fig:cyclic}
\end{figure}

\subsection{Count Generation}
\label{sec:count_gen}
Following UniLIP~\cite{tang2025unilip}, we adopt the SANA 
diffusion module~\cite{xie2025sana}, where the autoregressive LM 
head conditions the diffusion head via cross-attention. Directly 
using understanding encoder embeddings for generation scrambles 
spatial layout (\cref{fig:issues}), revealing a synergy gap from 
independent branch training, which we address next.

\noindent\textbf{Generation via Understanding} 
Since unified models like UniLIP~\cite{tang2025unilip} have 
disentangled autoregressive and diffusion heads with separate 
training objectives, a synergy gap emerges between the two branches. 
Yet the tasks are naturally complementary: the understanding branch 
can directly assess how well a generated image aligns with its 
conditioning prompt, providing internal feedback without external 
supervision. Rather than decoupling the two tasks in a 
``first understand, then generate'' pipeline, our approach embeds 
this feedback directly into generation: the understanding branch 
identifies failures in the generator's output (\eg, wrong counts 
or incorrect spatial arrangements) and guides the generation branch 
to correct them, transforming inter-branch antagonism into a 
catalyst for progressive generative improvement.

\begin{table*}[!t]
\centering
\setlength{\tabcolsep}{4pt}
\caption{Comprehensive evaluation of Object Counting (MAE $\downarrow$ / RMSE $\downarrow$) across FSC-147, CARPK, and ShanghaiTech (SHT), and Count Reasoning (EM $\uparrow$) on CountQA. $\dagger$ indicates methods using visual exemplars (few-shot). Methods are partitioned by supervision signal: Point-level (P), Image-level (I), and Zero-shot/MLLM (Z).}
\label{tab:countsota}
\scriptsize
\resizebox{\textwidth}{!}{
\begin{tabular}{llccccccccccc}
\toprule
& &
\multicolumn{6}{c}{\textbf{Object Counting}} &
\multicolumn{4}{c}{\textbf{Crowd Counting}} &
{\textbf{Reasoning}} \\
\cmidrule(lr){3-8}\cmidrule(lr){9-12}\cmidrule(lr){13-13}
\multirow{2}{*}{Method} & \multirow{2}{*}{Sup.} &
\multicolumn{2}{c}{FSC-147 Val} &
\multicolumn{2}{c}{FSC-147 Test} &
\multicolumn{2}{c}{CARPK Test} &
\multicolumn{2}{c}{SHT-A Test} &
\multicolumn{2}{c}{SHT-B Test} &
{CountQA} \\
\cmidrule(lr){3-4}\cmidrule(lr){5-6}\cmidrule(lr){7-8}\cmidrule(lr){9-10}\cmidrule(lr){11-12}\cmidrule(lr){13-13}
& & MAE & RMSE & MAE & RMSE & MAE & RMSE & MAE & RMSE & MAE & RMSE & EM (\%) $\uparrow$ \\
\midrule
\rowcolor{secbg}
\multicolumn{13}{l}{\textit{Specialist Counting Model}} \\
CountGD++~\cite{amini2025countgd++}               & P & 12.14 & 47.51 & 8.39  & 27.03  & --    & --    & 116.0 & 234.0  & 28.0  & 50.0  & -- \\
T2ICount~\cite{qian2025t2icount}                  & P & 13.78 & 58.78 & 11.76 & 97.86  & 8.61  & 13.47 & --    & --     & --    & --    & -- \\
CountSE$^{\dagger}$~\cite{liu2025countse}          & P & --    & --    & 7.84  & 82.99  & --    & --    & 129.7 & 258.3  & --    & --    & -- \\
CAD-GD~\cite{wang2025CADGD}                       & P & --    & --    & 10.35 & 86.88  & --    & --    & --    & --     & --    & --    & -- \\
\arrayrulecolor{ruleblue}\midrule\arrayrulecolor{black}
\rowcolor{secbg}
\multicolumn{13}{l}{\textit{VLM-based Counting Model}} \\
GPT-5.5~\cite{openai2026gpt55}                    & Z & 25.87 & 79.34  & 25.17 & 162.0  & {24.33} & {36.50} & {215.67} & {412.89} & {58.92} & {102.34} & 25.03 \\
Show-o~\cite{xie2024show}                         & Z & 37.87 & 105.55 & 46.26 & 129.53 & {42.15} & {63.22} & {312.45} & {587.33} & {89.34} & {156.78} & 7.85 \\
Janus Pro 7B~\cite{chen2025janus}                 & Z & 43.56 & 110.23 & 35.70 & 99.96  & {34.52} & {51.78} & {278.91} & {523.44} & {76.23} & {134.56} & 6.98 \\
UniLIP-3B~\cite{tang2025unilip}                   & Z & 30.19 & 103.07 & 26.44 & 103.98 & 26.87 & 33.48 & 243.15 & 424.33 & 63.79 & 97.04 & 9.23 \\
WS-COC-7B~\cite{zhang2026wscoc}                      & I & 14.77 & 54.24 & 13.91 & 97.28  & 10.39 & 15.83 & 128.9 & 232.9  & 34.2  & 57.0  & 8.44 \\
\midrule
\rowcolor{ourbg}
\textbf{\textsc{\model-3B}}                          & Z & \textbf{5.71} & \textbf{26.46} & \textbf{5.03} & \textbf{27.03} & \textbf{8.41} & \textbf{10.84} & \textbf{78.59} & \textbf{139.88} & \textbf{14.75} & \textbf{25.08} & \textbf{15.3} \\
\bottomrule
\end{tabular}
}
\end{table*}

\noindent\textbf{Count Aware Image Enhancement}
To close the understanding--generation synergy gap, we design a 
cycle-consistent self-supervised RL framework optimised via 
GRPO~\cite{shao2024deepseekmath} (in Eq~\ref{eq:grpo}). As illustrated in 
\cref{fig:cyclic}, given a count-conditioned prompt $\mathbf{t}$ 
specifying target cardinality $c^{*}$, the generation branch samples 
$K$ candidate images $\{\hat{\mathbf{x}}_i\}_{i=1}^{K} \sim 
\pi_\theta(\cdot \mid \mathbf{t})$. The frozen understanding branch 
$\overline{\mathcal{M}}_\theta$ then counts the target instances in 
each candidate:
\begin{equation}
    \hat{c}_i = \mathrm{parse}\!\left(
        \overline{\mathcal{M}}_\theta\!\left(
            \Pi(\mathcal{V}(\hat{\mathbf{x}}_i)),\, \mathbf{z}_t
        \right)
    \right),
    \label{eq:cycle_count}
\end{equation}
yielding a count-deviation reward consistent with \cref{eq:reward}:
\begin{equation}
    R_{\mathrm{cnt}}(\hat{\mathbf{x}}_i) =
        \exp\!\left(-\frac{|\hat{c}_i - c^{*}|}{c^{*} + \epsilon}
        \right).
    \label{eq:count_reward}
\end{equation}
To prevent reward hacking through cardinality correctness at the 
expense of visual quality — the failure mode of attention-manipulation 
baselines — we augment with an off-the-shelf aesthetic 
scorer~\cite{schuhmann2022laion} $S_{\mathrm{aes}}(\cdot)$, giving 
the composite reward:
\begin{equation}
    R(\hat{\mathbf{x}}_i) =
    \lambda_c\, R_{\mathrm{cnt}}(\hat{\mathbf{x}}_i)
    + \lambda_a\, S_{\mathrm{aes}}(\hat{\mathbf{x}}_i),
    \label{eq:gen_reward}
\end{equation}
where {$\lambda_c=1.0$ and $\lambda_a=0.1$} are fixed scalar weights. 
Group-relative advantages and the KL-regularised surrogate follow 
\cref{eq:advantage,eq:grpo}; gradients flow only into the 
generation-branch LoRA, with $\pi_{\mathrm{ref}}$ fixed to the 
SFT-initialised generator and the understanding branch frozen 
throughout. This asymmetry is essential: it prevents the count 
reward from corrupting the counter that produces it, while exposing 
the understanding branch to the evolving generator distribution 
at training time. As the generation branch improves, it produces increasingly 
realistic multi-instance scenes that progressively sharpen the 
understanding branch's reward signal, yielding emergent 
count-faithful synthesis that the generation branch could not 
achieve when trained in isolation.

\subsection{ABACUS Training Strategy}
\label{sec:train}
We adopt a three-stage training strategy to build a unified model 
for count understanding and generation.

\noindent\textbf{Stage 1: Understanding Branch Training.} We 
finetune the MLLM using density-aware zoomed images with the 
combined objective:
\begin{equation}
    \mathcal{L}_{\text{und}} = \mathcal{L}_{\text{SFT}} + 
    \mathcal{L}_{\text{obj}},
\end{equation}
where $\mathcal{L}_{\text{SFT}}$ is the next-token prediction loss 
and {$\mathcal{L}_{\mathrm{obj}}$} is the objectness regularisation 
from \cref{eq:focus_loss}. We train the MLLM, learnable query 
embeddings, and affine MLP on 2M densely annotated images via LoRA, 
enabling the model to output structured count predictions. We then 
apply post-training on 50K curated samples using 
$\mathcal{L}_{\text{GRPO}}$ to further reduce over/undercounting. 
The connector and generation branch remain frozen throughout.

\noindent\textbf{Stage 2: Connector Training.} 
{The primary focus of this training stage involves optimizing a connector module designed to interface the MLLM with the DiT. By keeping the parameters of both base models frozen and restricting training solely to generative tasks, the connector learns to accurately align the MLLM’s output representations with the DiT’s original conditional feature space. The connector is trained via SFT ( $\mathcal{L}_{SFT}$) with $1$M high-quality instruction tuning data respectively.
This connector acts as an architectural bridge to transfer the MLLM’s inherent world knowledge about count understanding, strong
reasoning, and in-context learning capabilities to image generation thus making it count-aware. We direcly feed a set of learnable queries into a frozen MLLM to extract multimodal conditions for multimodal generation which is then passed through this connector into the DiT branch. }

\noindent\textbf{Stage 3: Generation Branch Training.} We train 
both the connector and DiT on large-scale count-conditioned 
generation data, with the MLLM frozen following~\cite{tang2025unilip}. 
After SFT, the model can follow generation prompts but produces 
incorrect cardinalities. We therefore apply post-training via 
$\mathcal{L}_{\text{GRPO}}$ with combined count-deviation and 
aesthetic rewards (\cref{eq:gen_reward}), updating only the 
generation branch using count-conditioned prompts as the sole 
training signal to reduce the generation-understanding synergy.

%% file: final_arxiv/04_exp.tex
\section{Experiments}

\subsection{Training Data Collection}
Our 2M dense-annotated 
understanding dataset is curated from large-scale open-source 
sources including Objects365~\cite{shao2019objects365}, 
V3Det~\cite{wang2023v3det}, and SKU-110K~\cite{goldman2019dense}, 
retaining only images with a minimum instance count of five objects 
after numeracy checks, aesthetic filtering, and deduplication.
For generation and SFT, we collect 1M images from Pexels, sourced from 
web alt-text and captions, and filtered via CLIP-based similarity 
scoring, resolution and aspect-ratio constraints, and text-length 
checks. Concept-aware sampling is applied to mitigate long-tail 
distributions, and structured supervision from OCR, charts, and 
grounding annotations is included to strengthen spatial 
understanding. {These images are annotated with T-Rex2~\cite{jiang2023t} for category-labelled boxes, human-verified via T-Rex Label; we prompt Qwen2.5-VL-7B~\cite{Qwen2.5-VL} with image and verified categories to compose captions. Each prompt's count is the human-verified detector count — grounded, not VLM-estimated — ensuring training-data count accuracy.}
To ensure fair comparison, all training and test 
splits of FSC-147~\cite{ranjan2021learning}, 
CARPK~\cite{hsieh2017drone}, 
ShanghaiTech~\cite{zhang2016single}, 
REC~\cite{dai2024referring}, and 
CountQA~\cite{tamarapalli2025countqa} are strictly excluded from 
training.

\begin{table}[!h]
\centering
\setlength{\tabcolsep}{3.5pt}
\caption{Referring expression counting on REC-8K~\cite{dai2024rec}
test set ($n{=}3{,}153$ pairs). $\dagger$ uses exemplar images.
GroundingREC and finetuned GroundingDino are specialist detection-based
methods trained with box-level supervision on REC-8K;
\model~is a unified VLM evaluated text-only with no
{training on evaluation benchmark data}.}
\label{tab:rec}
\small
\begin{tabular}{llccc}
\toprule
Method & Backbone & FT & MAE\,$\downarrow$ & RMSE\,$\downarrow$ \\
\midrule
\rowcolor{secbg}
\multicolumn{5}{l}{\textit{Specialist Counting Model}} \\
ZSC~\cite{xu2023zsc}                          & Swin-T   & \cmark  & 13.00            & 29.07 \\
TFOC~\cite{shi2024tfoc}                       & ViT-B    & --      & 17.27            & 32.68 \\
CounTX~\cite{amini2023countx}                 & ViT-B/16 & \cmark  & 11.84            & 25.62 \\
GDino~\cite{liu2024grounding}         & Swin-T   & \cmark  & \phantom{0}8.88  & 21.95 \\
GrREC~\cite{dai2024rec}                & Swin-T   & \cmark  & \phantom{0}\textbf{6.50} & 19.79 \\
\arrayrulecolor{ruleblue}\midrule\arrayrulecolor{black}
\rowcolor{secbg}
\multicolumn{5}{l}{\textit{Baselines}} \\
UniLIP-3B \cite{tang2025unilip}                                 & --       & --      & 13.75            & 25.91 \\
\midrule
\rowcolor{ourbg}
\textbf{\model-3B}                               & UniLIP-3B & --     & \phantom{0}7.67  & \textbf{15.84} \\
\bottomrule
\end{tabular}
\end{table}

\subsection{Evaluation Metrics}

Following common practice in object and crowd
counting~\cite{ranjan2021learning,zhang2016sHAB}, we report
Mean Absolute Error (MAE) and, where standard, Root Mean Squared Error (RMSE).
For count generation we follow Make-It-Count~\cite{binyamin2025make} and
report YOLOv9~\cite{wang2024yolov9} exact-match accuracy on
CoCoCount~\cite{binyamin2025make}, the Numeracy score on
T2I-CompBench~\cite{binyamin2025make}, the Counting subtask of
GenEval~\cite{bakhtingeneval} averaged over multiple seeds.
For referring-expression counting we follow the protocol of
GroundingREC~\cite{dai2024rec} and report MAE on REC-8K.

\subsection{Implementation Details}

{\myparagraph{Model Architecture and LoRA Configuraion} \model{} builds on UniLIP-3B~\cite{tang2025unilip} (InternViT encoder, Qwen2 LLM, SANA-1.5B DiT, DC-AE pixel decoder) connected via a projector $\Pi$ and $N{=}256$ learnable queries. We apply LoRA ($r{=}32, \alpha{=}64$) to all Qwen2 attention ($W_Q, W_K, W_V, W_O$) and FFN ($W_{\text{up}}, W_{\text{down}}$) projections, adding ${\sim}$48M trainable parameters (1.6\% overhead). The multimodal projector $\Pi$ and $W_{\texttt{lm\_head}}$ are fully trainable. For text-to-image generation, a separate LoRA ($r{=}16, \alpha{=}32$) is applied to the SANA-1.5B DiT. The visual encoder $\mathcal{V}$, pixel decoder, and query bank remain strictly frozen.  The connector consists of $6$ layers and maintains the same structure and dimensions as the LLM architecture in InternVL3\cite{chen2024internvl}. 
%
%
\\ \myparagraph{Post-Training (GRPO)} 
\textit{Boundary Policy} is trained on 8K dense images for 2K steps (LR $5{\times}10^{-6}$). We sample $K{=}4$ rollouts per image with a strong KL penalty ($\beta{=}0.04$) to stabilize the structured spatial rewards.
\textit{Cycle-Consistent Generation} is trained for 5K steps (LR $1{\times}10^{-6}$). The frozen understanding branch scores 8 DiT candidates per prompt using reward $r = \exp(-|\hat{c} - c^{*}|/c^{*})$. We use a relaxed $\beta{=}0.01$ to allow synthesis freedom. At inference, this Best-of-8 strategy yields 80\% exact numeracy.
\\ \myparagraph{Density Indicator \& Compute} We use a frozen GroundingDINO-T with a 2-layer MLP classifier trained on 15K sparse/dense images. At inference, regions scoring $\geq 0.5$ trigger recursive $2{\times}2$ partitioning (max depth $\gamma{=}3$). The entire pipeline is trained in bfloat16 mixed precision, taking ${\sim}$44 hours on $8{\times}$ NVIDIA A100 80GB GPUs.}

\subsection{Comparison with State-of-the-art methods}

\change{\myparagraph{Object and Crowd Counting}
Tab~\ref{tab:countsota} compares \model{} against specialist and 
VLM-based counters. 
On FSC-147, \model{} achieves 5.71 val MAE and 5.03 test MAE, 
surpassing the strongest specialist (CountGD++, 12.14/8.39) by 
over 40\% without point-level annotations. 
On CARPK, \model{} attains 8.41 MAE, outperforming the only 
reporting specialist T2ICount (8.61). 
On crowd counting, \model{} achieves 78.59/14.75 MAE on 
ShanghaiTech-A/B, roughly halving the error of both the best 
specialist (CountGD++: 116.0/28.0) and the best VLM-based method 
(WS-COC-7B: 128.9/34.2). 
Among VLM-based methods, gains over the base UniLIP-3B are 
$5\times$ on FSC-147 val and $3\times$ on CARPK, confirming that 
the counting adapter, objectness map, and boundary-aware policy 
together close the gap to task-specific specialists. 
Notably, \model{} is the only method that simultaneously achieves 
state-of-the-art on both object and crowd counting from a single 
model. 
On the count reasoning benchmark 
CountQA~\cite{tamarapalli2025countqa}, \model{} achieves 15.3\% 
EM, the highest among open unified VLMs of comparable scale, 
surpassing UniLIP-3B (9.23\%), WS-COC-7B (8.44\%), Janus Pro 7B 
(6.98\%), and Show-o (7.85\%), respectively.}

\begin{table}[!t]
\centering
\setlength{\tabcolsep}{2.5pt}
\caption{Count Generation evaluation across CoCoCount, T2I-CompBench, and GenEval. YOLOv9 ($\uparrow$) for CoCoCount and GenEval; Human count accuracy ($\uparrow$) from annotator study (see supple.); Aesthetic Quality ($\uparrow$) on GenEval (per-benchmark aesthetics in supplementary).}
\label{tab:gensota}
\scriptsize
\resizebox{\columnwidth}{!}{
\begin{tabular}{lcccccc}
\toprule
& \multicolumn{2}{c}{CoCoCount} &
{T2I-Comp} &
\multicolumn{3}{c}{GenEval} \\
\cmidrule(lr){2-3}\cmidrule(lr){4-4}\cmidrule(lr){5-7}
Method
& YOLOv9 & Hum.
& Hum.
& YOLOv9 & Hum. & Aes. \\
\midrule
\rowcolor{secbg}
\multicolumn{7}{l}{\textit{Specialist Count Generation}} \\
CountGen~\cite{binyamin2025make}       & 50 & 54 & 48 & 46 & 44 & 45 \\
BoundedAttn~\cite{dahary2024yourself}  & 29 & 30 & 35 & 21 & 18 & 10 \\
Count.\ Guid.~\cite{kang2023countingguidance} & 21 & 22 & 22 & 16 & 11 & 7 \\
\arrayrulecolor{ruleblue}\midrule\arrayrulecolor{black}
\rowcolor{secbg}
\multicolumn{7}{l}{\textit{VLM-based Count Generation}} \\
BAGEL~\cite{deng2025bagel}             & 36 & 41 & 32 & 44 & 39 & 43 \\
Janus Pro-7B~\cite{chen2025janus}      & 27 & 32 & 25 & 30 & 33 & 58 \\
UniLIP-3B~\cite{tang2025unilip}        & 34 & 39 & 30 & 36 & 40 & 61 \\
\midrule
\rowcolor{ourbg}
\textbf{\textsc{\model}}               & \textbf{71} & \textbf{77} & \textbf{65} & \textbf{94} & \textbf{95} & \textbf{89} \\
\bottomrule
\end{tabular}
}
\end{table}

\myparagraph{Referring Expression Counting}
On REC-8K~\cite{dai2024rec}, \model{} is queried with free-form 
referring expressions (\eg~``\textit{red apples on the left}'') 
without any architectural modification, achieving MAE~7.67 and 
RMSE~15.84 over 3{,}153 evaluation pairs. \model{} surpasses the 
fine-tuned GDINO~\cite{liu2023grounding} specialist and 
matches GrREC~\cite{dai2024rec}, a detection-trained 
model with box-level supervision, while achieving substantially 
lower RMSE (15.84 vs.\ 19.79). To our knowledge, this is the 
strongest text-only unified-VLM result on this benchmark.

\myparagraph{Count Image Generation}
Tab~\ref{tab:gensota} evaluates count-faithful generation on 
CoCoCount, T2I-CompBench, and GenEval. \model{} achieves 71\% 
YOLOv9 exact-match on CoCoCount, surpassing the previous best 
(CountGen, 50\%) by 21 points, and 94 on GenEval counting versus 
46 for CountGen. Among unified VLMs, \model{} nearly doubles the 
accuracy of the closest competitor BAGEL (36\%) with a smaller 3B 
backbone. Beyond count accuracy, \model{} achieves the highest 
aesthetic quality on GenEval (89 vs.\ 61 for UniLIP-3B), while 
specialist methods (BoundedAttn, Counting Guidance) score only 
7--10 due to attention manipulation and  degrading image coherence. 
Full human evaluation results are in the supplementary, where 
\model{} is preferred 39\%, 41\%, and 50\% of the time on 
CoCoCount, T2I-CompBench, and GenEval respectively, far exceeding 
the 20\% random baseline. {An interesting observation of our model is that,  during generation branch training, ABACUS relies on the frozen understanding branch, which counts via visible object centers (from objectness regularisation); hence the generation branch thus favours visible centers even under occlusion which we can observe visually in Fig~\ref{fig:gen}(row 5) where our model can generate 28 "colorful balls" with occlusion including the balls underneath which makes it a more realistic generation.}

\subsection{Ablation Study}
\label{sec:ablation}

We ablate the core components of \model{} on FSC-147 val (MAE/RMSE) for understanding and CoCoCount (YOLOv9 exact-match) for generation. All variants share the same LoRA adapter, training data, and hyperparameters unless stated otherwise.

\myparagraph{Objectness map}
We ablate the MHSA-derived objectness map by comparing the full 
pipeline against: (a)~a naive mean-pooled attention map across all 
heads and layers, and (b)~removing objectness regularisation 
entirely ($\mathcal{L}_{\text{obj}}{=}0$). As shown in 
\cref{tab:abl_objectness}, removing $\mathcal{L}_{\text{obj}}$ 
causes the largest degradation (+3.92 MAE). Partitioning FSC-147 
val into overlap-heavy (${\geq}20\%$ of GT points within 8\,px) 
and overlap-light subsets, the MAE gap between subsets is 2.31 
with the full model but widens to 6.18 without 
$\mathcal{L}_{\text{obj}}$, confirming that the objectness map 
primarily helps disambiguate spatially proximate instances.

\begin{table}[!h]
\centering
\caption{Ablation of the objectness map on FSC-147 val.}
\label{tab:abl_objectness}
\footnotesize
\begin{tabular}{lcc}
\toprule
\textbf{Variant} & \textbf{MAE} $\downarrow$ & \textbf{RMSE} $\downarrow$ \\
\midrule
Full (per-head + $\mathcal{A}$ + $\mathcal{L}_{\mathrm{obj}}$) & \textbf{5.71} & \textbf{26.46} \\
Mean-pooled attention & 7.94 & 34.21 \\
No $\mathcal{L}_{\mathrm{obj}}$ & 9.63 & 40.15 \\
\bottomrule
\end{tabular}
\end{table}

\myparagraph{Density-aware adaptive zooming}
We ablate the zooming module $\phi(\cdot)$ (~\cref{eq:zoom}): (a)~{no zooming} (single-pass), (b)~{fixed $2{\times}2$ grid} (unconditional split), and {~{adaptive} partitioning (comparing the the objectness map(c) against our G-DINO density indicator(d))}
As shown in \cref{tab:abl_zoom}, single-pass inference suffers on dense scenes.
Fixed-grid partitioning improves over single-pass but introduces
double-counting at tile boundaries on sparse images, the failure
mode that motivates the boundary-aware count policy, ablated in
the supplementary. Adaptive zooming avoids both failure modes, yielding the best MAE within $1.2-1.3{\times}$ of single-pass inference.
{Furthermore, relying on the objectness map instead of a dedicated G-DINO indicator is unreliable for density partitioning, yielding a higher MAE of 8.64 compared to 5.71 for our approach.}

\begin{table}[h]
\centering
\caption{Ablation of density-aware zooming and indicator choice on FSC-147 val.}
\label{tab:abl_zoom}
\footnotesize
\begin{tabular}{lccc}
\toprule
\textbf{Variant} & \textbf{MAE} $\downarrow$ & \textbf{RMSE} $\downarrow$ & \textbf{Rel.\ Time} \\
\midrule
No zooming (single pass) & 10.87 & 45.32 & $1.0\times$ \\
Fixed $2{\times}2$ grid & 7.93 & 34.58 & ${\sim}1.8\times$ \\
{Adaptive (Objectness Map)} & {8.64} & {37.21} & {${\sim}1.3\times$} \\
Adaptive (G-DINO indicator) & \textbf{5.71} & \textbf{26.46} & ${\sim}1.2\times$ \\
\bottomrule
\end{tabular}
\end{table}

\myparagraph{Count-aware image enhancement}
The generation branch is optimised via cycle-consistent GRPO (\cref{sec:count_gen}): generate $\to$ self-count $\to$ compare to prompt $\to$ update generator.
We compare (\cref{tab:abl_gen_grpo}): (a)~{no GRPO} (LoRA SFT only), (b)~{open-loop GRPO} (frozen external counter as reward source), and (c)~the full {cycle-consistent GRPO}.
SFT alone achieves only 45\% exact-match.
Open-loop GRPO improves by +17 points, but the full cycle outperforms it by a further 9 points, confirming that co-adaptation of both branches, where the understanding head sharpens {the} generated images, producing progressively more informative rewards is itself a meaningful source of gain.

\begin{table}[!h]
\centering
\caption{Ablation of count-aware image enhancement on CoCoCount.}
\label{tab:abl_gen_grpo}
\footnotesize
\begin{tabular}{lc}
\toprule
\textbf{Variant} & \textbf{CoCoCount (exact-match $\uparrow$)} \\
\midrule
No GRPO (LoRA SFT only) & 45 \\
Open-loop GRPO (ext.\ counter) & 62 \\
Full cycle-consistent GRPO & \textbf{71} \\
\bottomrule
\end{tabular}
\end{table}

\myparagraph{Choice of counting head:} We use the same \model{} model to count objects via two readouts:
(i)~\emph{objectness peak}, where the binarised objectness map
$\tilde{q}^l(v)$ is thresholded and connected components are counted
directly, and
(ii)~\emph{autoregressive}, where the language head generates the count
as text tokens.
\model{} uses the autoregressive readout by default. We compare both
from the same model (\cref{tab:supp_abl_head}) with no additional
parameters or training.
Objectness peak counting is competitive on dense images (${\geq}50$ GT
objects), where spatial peaks are well-separated, but degrades on sparse
images ($<10$ GT objects) where the $14{\times}14$ patch resolution
merges nearby instances and isolated peaks are sensitive to the
binarisation threshold.
The autoregressive readout outperforms peak counting across both
regimes, confirming that $\mathcal{L}_{\mathrm{focus}}$ successfully
internalises spatial structure into the language model's hidden states
during training, making the explicit spatial readout redundant at
inference.

\begin{table}[!h]
\centering
\caption{Counting using autoregressive head vs.\ objectness peak
on FSC-147 val. Sparse: $<$10 GT; Dense: $\geq$50 GT.}
\label{tab:supp_abl_head}
\small
\setlength{\tabcolsep}{3.5pt}
\begin{tabular}{lcccc}
\toprule
& \multicolumn{2}{c}{Overall} & {Sparse} & {Dense} \\
\cmidrule(lr){2-3}\cmidrule(lr){4-4}\cmidrule(lr){5-5}
Readout & MAE\,$\downarrow$ & RMSE\,$\downarrow$ &
  MAE\,$\downarrow$ & MAE\,$\downarrow$ \\
\midrule
Objectness peak counting
  & 8.52  & 36.71 & 6.31  & 12.14 \\
Autoregressive (\model)
  & \textbf{5.71} & \textbf{26.46} & \textbf{3.42} & \textbf{9.87} \\
\bottomrule
\end{tabular}
\end{table}

\begin{figure}[!h]
  \centering
  \includegraphics[width=\columnwidth]{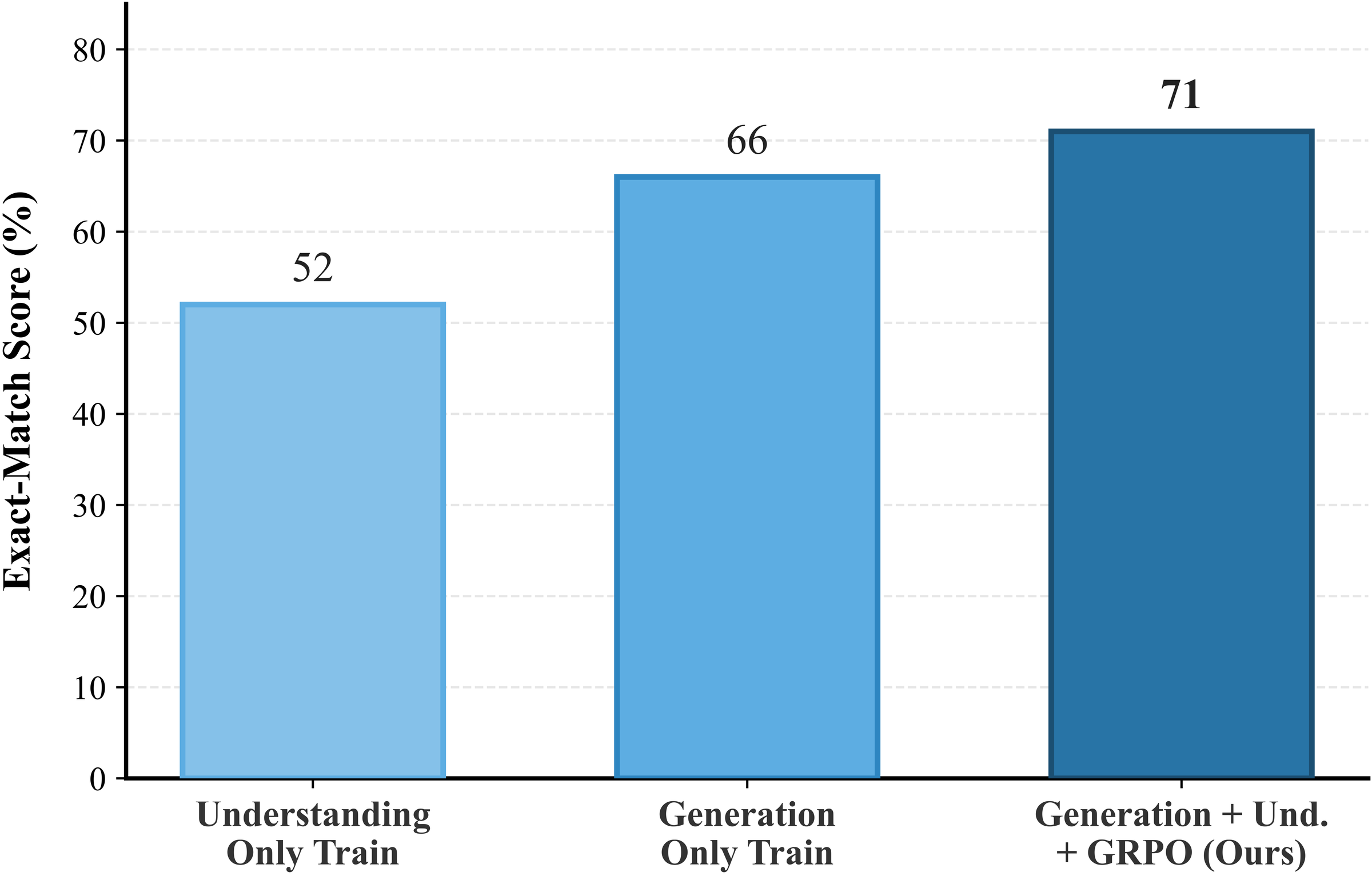}
  \caption{{Ablation of Training Strategy on CoCoCount.}}
  \label{fig:abl_training}
\end{figure}

\myparagraph{Training strategies}
{ABACUS follows a staged initialization plus cycle-consistent GRPO coupling, not a single joint objective. The process fine-tunes UniLIP-3B weights: (i) fine-tune the understanding branch; (ii) freeze it, fine-tune the connector and generation branch on numeracy prompt/image pairs; (iii) GRPO post-training updating only the generation branch using the frozen understanding branch as a reward. As shown in \cref{fig:abl_training}, this staged coupling yields complementary gains over either branch trained alone.}


\myparagraph{Backbone generalisability}
\change{
To verify that \model{} is not specific to UniLIP-3B, we apply 
the full adapter pipeline to BAGEL-7B~\cite{deng2025bagel} and 
Nexus-Gen-7B~\cite{zhang2025nexus}. As shown in 
\cref{tab:abl_backbone}, \model{} yields consistent gains across 
all backbones, with performance scaling with model capacity: 
BAGEL-7B + \model{} achieves the lowest FSC-147 val MAE (4.93) 
and highest CoCoCount exact-match (76). This confirms that the 
adapter pipeline is architecture-agnostic and that scaling the 
backbone translates directly into stronger counting and generation. 
All main results are reported on UniLIP-3B to demonstrate 
state-of-the-art performance at the 3B scale.
}
 
\begin{table}[h]
\centering
\caption{Backbone generalisability of the \model{} adapter.
Larger backbones yield stronger results.}
\label{tab:abl_backbone}
\footnotesize
\begin{tabular}{lcccc}
\toprule
& \multicolumn{2}{c}{FSC-147 Val MAE\,$\downarrow$} &
  \multicolumn{2}{c}{CoCoCount\,$\uparrow$} \\
\cmidrule(lr){2-3}\cmidrule(lr){4-5}
Backbone & Base & +\,\model & Base & +\,\model \\
\midrule
UniLIP-3B     & 30.19 & 5.71 & 34 & 71 \\
Nexus-Gen-7B  & 34.81 & 5.28 & 32 & 73 \\
BAGEL-7B      & 32.45 & \textbf{4.93} & 36 & \textbf{76} \\
\bottomrule
\end{tabular}
\end{table}

\myparagraph{High cardinality generation limits}
{Exact matches remain challenging at extreme densities. As shown in \cref{tab:high_cardinality}, quality degrades as the requested instance count increases, a physical packing limitation inherent to diffusion models. Nonetheless, ABACUS offers lower error than task-specific specialists across regimes and operates as a unified open-vocabulary model without task-specific training, unlike density regressors.}

\begin{table}[h]
\centering
{
\caption{{Count Accuracy and Aesthetics vs. Requested Cardinality.}}
\label{tab:high_cardinality}
\footnotesize
\begin{tabular}{lcc}
\toprule
\textbf{Count} & \textbf{Accuracy} & \textbf{Aesthetics} \\
\midrule
10 & 72.4\% & 0.91 \\
50 & 71.1\% & 0.72 \\
100 & 68.3\% & 0.65 \\
\bottomrule
\end{tabular}
}
\end{table}

\begin{figure}[!h]
  \centering
  \includegraphics[width=\columnwidth]{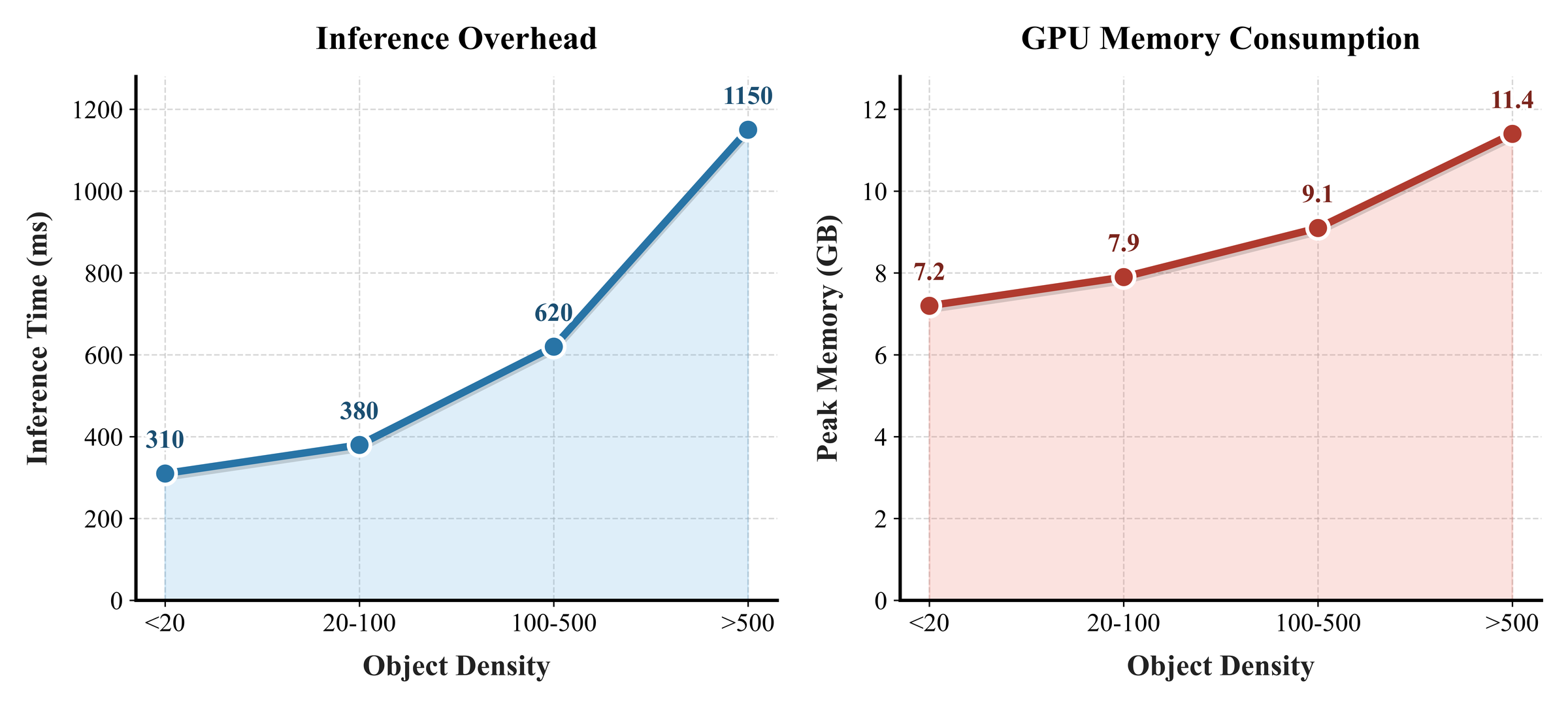}
  \caption{\change{Inference cost in dense scenes.}}
  \label{fig:inference_cost}
\end{figure}

\myparagraph{Inference cost on dense scenes}
Adaptive zooming keeps average inference time within $1.2{\times}$
of single-pass by partitioning only when the density indicator
triggers.
For extremely dense scenes requiring maximum recursion depth
$\gamma$, worst-case latency grows with the number of sub-regions.
In practice, such images are rare in standard benchmarks ($<$3\% of
FSC-147 val), and the accuracy gains substantially outweigh the
cost; nevertheless, early-exit strategies or learned depth
prediction could further optimise the speed-accuracy trade-off.
{We provide a detailed breakdown of inference latency and GPU memory overhead across different density regimes in \cref{fig:inference_cost}. The overhead is minimal for sparse scenes and scales gracefully up to extremely dense scenes.}


%% file: final_arxiv/05_conc.tex
\section{Limitations and future work}
 
\begin{figure}[!h]
  \centering
  \includegraphics[width=\columnwidth]{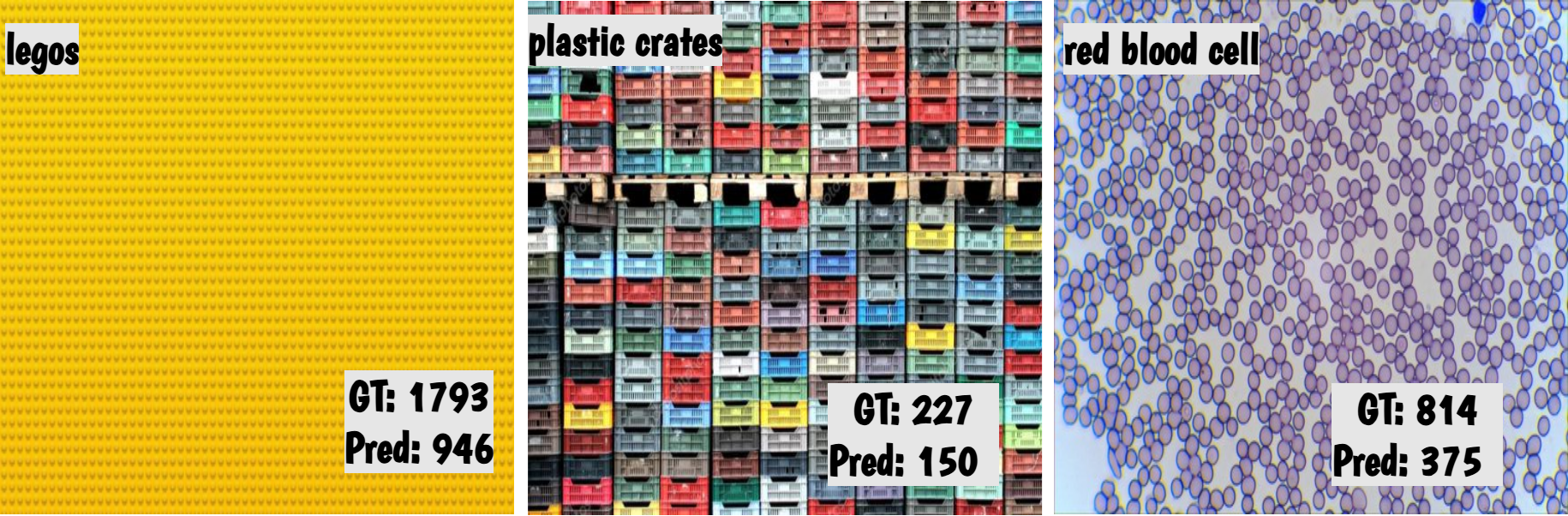}
  \caption{\change{Some failure cases of \model.}}
  \label{fig:lim}
\end{figure}

\myparagraph{Low-resolution and degraded inputs}
\model's counting pipeline relies on spatial tokens from InternViT's
$14{\times}14$ patch encoding, which requires sufficient input
resolution to distinguish individual instances.
On low-resolution ($<$224\,px) or heavily compressed images, common
in surveillance and legacy datasets, where the visual token grid becomes
too coarse for the objectness map to resolve individual objects,
a limitation shared by all patch-based VLMs and detection-based
counters alike~\cite{amini2025countgd++} (see \cref{fig:lim}.
Incorporating resolution-adaptive visual encoders or
super-resolution preprocessing could extend \model{} to these
degraded settings.

\myparagraph{Single-model generality vs.\ domain adaptation}
\change{A single 3B-parameter model achieves state-of-the-art results
across seven benchmarks spanning diverse visual domains.
Specialised domains such as medical cell counting~\cite{xie2018microscopy} (see \cref{fig:lim}),
satellite vehicle detection, or industrial defect inspection present
distribution shifts that the current count-balanced training mixture
does not cover.
Lightweight domain adaptation of the LoRA adapter, without
retraining the frozen backbone, could unlock these verticals while
preserving the general-purpose counting ability, and we leave this
exploration to future work.}

\section{Conclusion}
We presented \model{}, a unified vision-language model for 
count-aware image understanding and count-faithful generation 
within a single architecture. Our approach rests on three 
contributions: an objectness map from MHSA head decomposition 
that spatially grounds count predictions using point
supervision; a novel boundary-aware count policy trained via GRPO with 
nested rewards that eliminates over/undercounting at crop 
boundaries; and a cycle-consistent self-reward strategy where the 
understanding branch counts objects in the generator's own output, 
closing the feedback loop that external-critic methods leave open. 
With a single 3B-parameter model, \model{} sets a new state 
of the art across object counting, crowd counting, 
referring-expression counting, and count-faithful generation, 
surpassing both task-specific specialists and larger generalist 
models, while establishing the first unified-VLM result on 
CountQA. These results support a broader justification: count 
understanding and generation are not competing objectives but 
mutually reinforcing tasks whose joint optimisation yields 
emergent spatial awareness that neither specialist alone can 
achieve, suggesting that the cycle-consistent self-reward paradigm 
can extend beyond counting to other spatial reasoning tasks where 
the same model both produces and verifies its outputs.

\begin{acks}
The authors gratefully acknowledge NVIDIA Corporation for support through the NVIDIA Academic Grant Program, which provided computational resources for this research. The authors also acknowledge the use of resources provided by the Isambard-AI National AI Research Resource (AIRR) \cite{mcintoshsmith2024isambardai}, operated by the University of Bristol and funded by the UK Government’s Department for Science, Innovation and Technology (DSIT) through UK Research and Innovation and the Science and Technology Facilities Council [ST/AIRR/I-A-I/1023]. The idea of this paper was formulated while Sauradip was at University of Surrey, UK.
\end{acks}

\newpage
\begin{figure*} [t]
    \centering
    \includegraphics[width=0.86\linewidth]{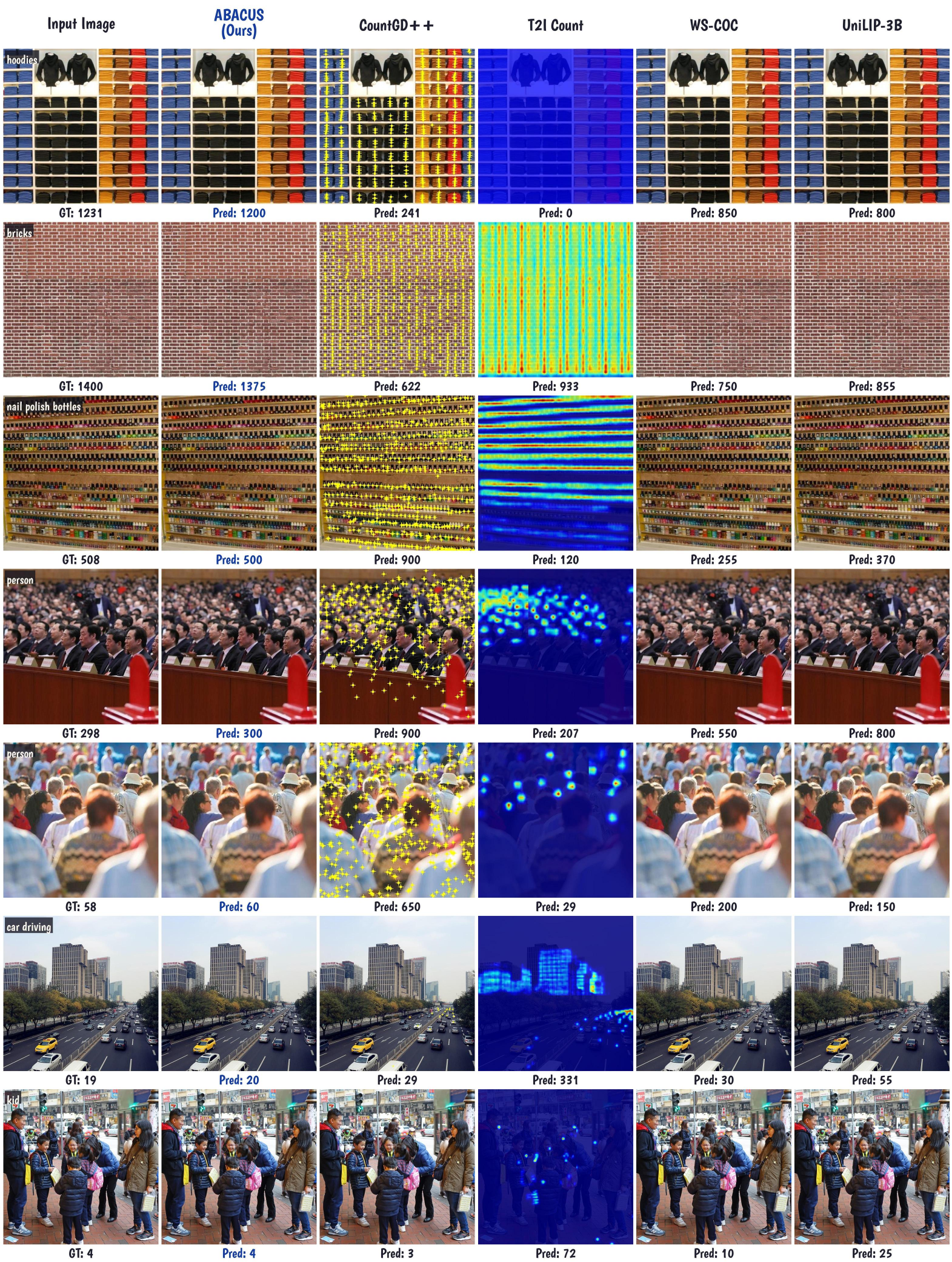}
    \caption{\change{\textbf{{Qualitative comparison on count understanding.}} \model{} (Ours) tracks the ground truth across all density regimes, from sparse (GT:~4) to extremely dense (GT:~1400). CountGD++ overcounts in dense scenes (900 for GT:~298); T2I~Count catastrophically fails on out-of-distribution layouts (0 for GT:~1231); WS-COC and UniLIP-3B default to coarse magnitude estimates. {Note: In cases of occlusion, counting relies on visible object centers, preventing overcounting from obscured parts.}}}
    \label{fig:und}
\end{figure*}
\newpage
\begin{figure*}
    \centering
    \includegraphics[width=0.81\linewidth]{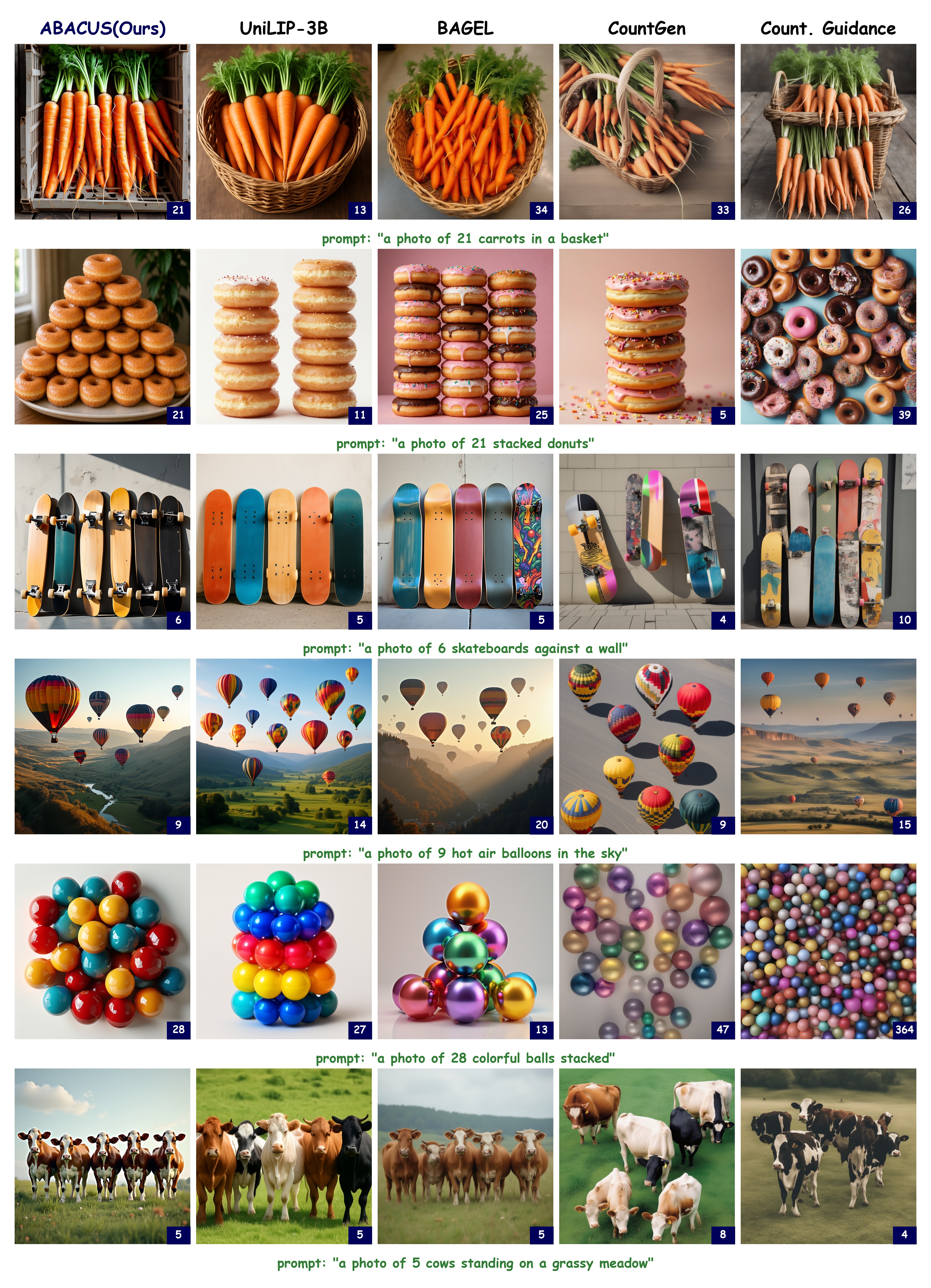}    \caption{\change{\textbf{Qualitative comparison on count generation.} \model{} achieves exact or near-exact counts while preserving natural spatial arrangement. UniLIP-3B systematically undercounts; BAGEL overcounts with unnatural compositions; CountGen produces rigid grid patterns; Counting Guidance exhibits mode collapse ({39} donuts for a prompt of {21}, 364 balls for a prompt of 28).}}
    \label{fig:gen}
\end{figure*}